\documentclass[sigconf]{acmart}
\usepackage[latin9]{inputenc}
\usepackage{array}
\usepackage{verbatim}
\usepackage{float}
\usepackage{url}
\usepackage{enumitem}
\usepackage{multirow}
\usepackage{amsmath}
\usepackage{graphicx}

\makeatletter

\providecommand{\tabularnewline}{\\}

\usepackage{booktabs}
\usepackage{stfloats}

\setcopyright{acmcopyright}
\usepackage{array}
\usepackage{mathtools}

\acmISBN{978-1-4503-4906-2/17/10}

\acmConference{MM '17}{October 23--27, 2017}{Mountain View, CA,
USA}\acmPrice{15.00}\acmDOI{10.1145/3123266.3123425}
\acmYear{2017}
\copyrightyear{2017}

\acmPrice{15.00}
\DeclareMathOperator*{\T}{\scriptscriptstyle \top}

\makeatother

\begin{document}
\title{Laplacian-Steered Neural Style Transfer}
\author{Shaohua Li$^{1,2}$ \hspace{3em} Xinxing Xu$^{2}$ \hspace{5em} Liqiang Nie$^{3}$ \hspace{4em} Tat-Seng Chua$^{1}$ \\     \hspace{-1em}\normalsize{shaohua@gmail.com} \hspace{1.5em} \normalsize{xuxinx@ihpc.a-star.edu.sg} \hspace{4em} \normalsize{nieliqiang@gmail.com} \hspace{4.5em} \normalsize{dcscts@nus.edu.sg} \\          \normalsize{1. School of Computing, National University of Singapore} \\  \normalsize{2. Institute of High Performance Computing, Singapore} \hspace{5em} \normalsize{3. Shandong University}}
\renewcommand{\shortauthors}{S. Li et al.}
\begin{abstract}
Neural Style Transfer based on Convolutional Neural Networks (CNN)
aims to synthesize a new image that retains the high-level structure
of a content image, rendered in the low-level texture of a style image.
This is achieved by constraining the new image to have high-level
CNN features similar to the content image, and lower-level CNN features
similar to the style image. However in the traditional optimization
objective, low-level features of the content image are absent, and
the low-level features of the style image dominate the low-level detail
structures of the new image. Hence in the synthesized image, many
details of the content image are lost, and a lot of inconsistent and
unpleasing artifacts appear. As a remedy, we propose to steer image
synthesis with a novel loss function: the Laplacian loss. The Laplacian
matrix (``Laplacian'' in short), produced by a Laplacian operator,
is widely used in computer vision to detect edges and contours. The
Laplacian loss measures the difference of the Laplacians, and correspondingly
the difference of the detail structures, between the content image
and a new image. It is flexible and compatible with the traditional
style transfer constraints. By incorporating the Laplacian loss, we
obtain a new optimization objective for neural style transfer named
Lapstyle. Minimizing this objective will produce a stylized image
that better preserves the detail structures of the content image and
eliminates the artifacts. Experiments show that Lapstyle produces
more appealing stylized images with less artifacts, without compromising
their ``stylishness''.
\end{abstract}
\maketitle

\section{Introduction}

\begin{figure}
\begin{centering}
\hspace*{-0.2cm}%
\begin{tabular}{>{\centering\arraybackslash}m{8em}>{\centering\arraybackslash}m{9em}>{\centering\arraybackslash}m{9em}}
\includegraphics[scale=0.128]{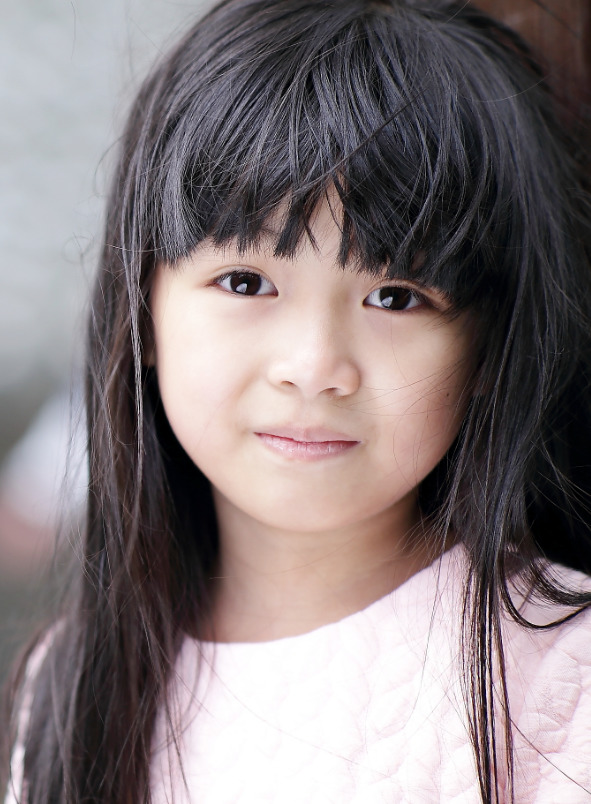} & \includegraphics[scale=0.12]{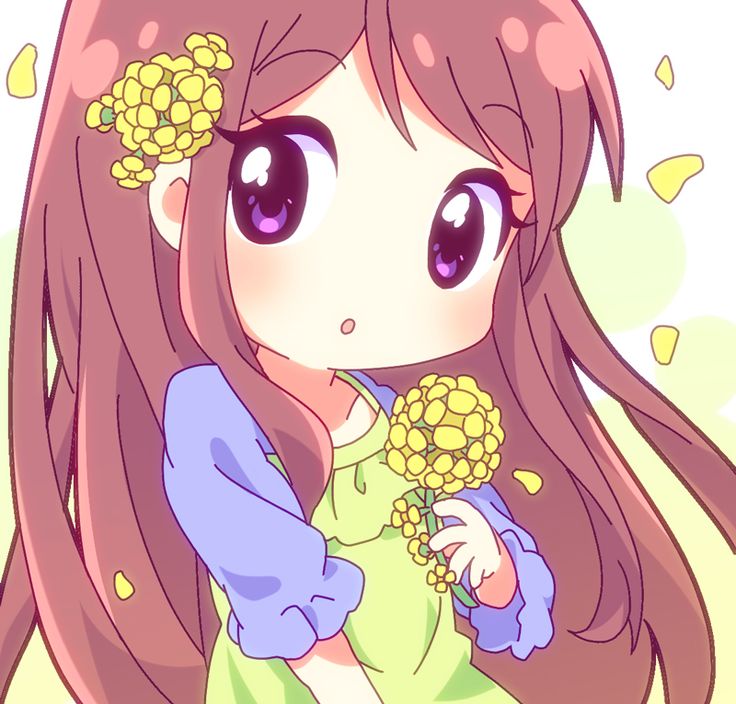} & \includegraphics[scale=0.2]{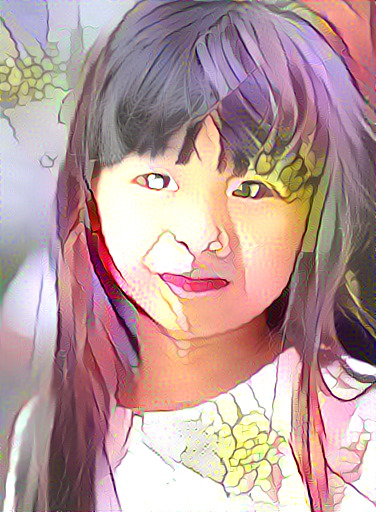}\tabularnewline
(a) Content image & (b) Style image & (c) Gatys-style\tabularnewline
\end{tabular}
\par\end{centering}
\caption{\label{fig:girlexample}Given a content image (a) and a style image
(b), style transfer creates a new one combining the content of (a)
and the style of (b). (c): the stylized image by Gatys et al. \cite{gatys16}
contains a lot of unpleasing artifacts, such as those on the girl's
face.}
\end{figure}

Neural Style Transfer based on Convolutional Neural Networks (CNNs)
\cite{gatys16,faststyle,gatys17,mrf} has gained a lot of attention
from both academia and industry. Given two input images, a content
image and a style image, the Neural Style Transfer algorithm synthesizes
a new image, referred to as the \emph{stylized image}, which retains
the basic structure and semantic content of the content image, but
is rendered in similar texture as the style image. Fig. \ref{fig:girlexample}
presents an example of style transfer: given a content image \ref{fig:girlexample}(a)
and a style image \ref{fig:girlexample}(b), a stylized image \ref{fig:girlexample}(c)
is synthesized. This combination is achieved thanks to the factorized
representations of an image by CNNs: the high-level CNN features compactly
encode the semantic content, e.g. the objects and global structure
of the input image, whereas the low-level features encode the local
details, e.g. colors, basic shapes and texture. Neural Style Transfer
specifies a total loss for a new image, which consists of both the
content loss, i.e. its high-level feature difference with the content
image, and the style loss, i.e. its low-level feature difference with
the style image. An image minimizing the total loss will combine the
global semantic content of the content image and the local texture
of the style image. One limitation of the current methods is that,
they do not incorporate the local features of the content image into
the optimization objective. Consequently, the \emph{detail structures}
(edges, contour segments, patterns, gradual color transitions, etc.)
of the stylized image are dominated by the low-level style image features.
After the optimization, visual elements in the style image may be
inappropriately transferred to random positions in the stylized image,
forming unappealing irregular artifacts, and some detail structures
in the content image may be severely distorted. These artifacts and
distortions are especially prominent on regions such as faces or object
contours, to which human perception are sensitive (e.g. Fig. \ref{fig:girlexample}c).
For notational convenience, the traditional method by Gatys et al.
\cite{gatys16} is referred to as \emph{Gatys-style}.

Naturally, to improve the structural coherence of neural style transfer,
more constraints with respect to the content detail structures could
be incorporated. When adding new content constraints, however, one
need to be cautious not to disrupt the optimization of the style loss.
A rigid content constraint will greatly reduce the search space for
feasible solutions, where a solution with a small style loss may not
exist. For example, in Section \ref{sec:Experiments}, we empirically
show that a constraint w.r.t. some low-level CNN features of the content
image is too rigid and greatly reduces the ``stylishness'' of the
result image.

\begin{figure}
\noindent \centering{}\hspace*{-0.5cm}%
\begin{tabular}{ccc}
\includegraphics[scale=0.127]{images/girl2} & \includegraphics[scale=0.262]{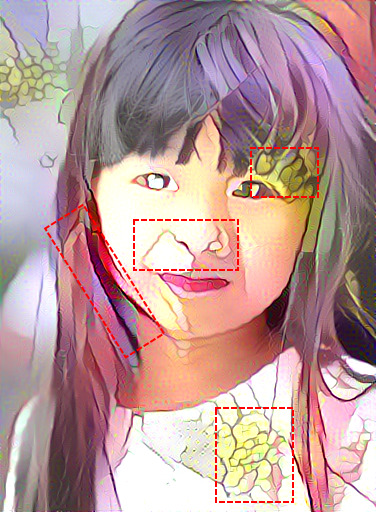} & \includegraphics[scale=0.252]{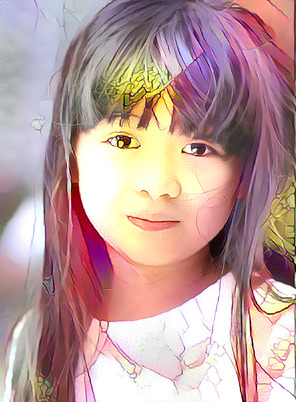}\tabularnewline
{\small{}(a) Content image} & {\small{}(b) Gatys-style output} & {\small{}(c) Lapstyle output}\tabularnewline
\includegraphics[scale=0.248]{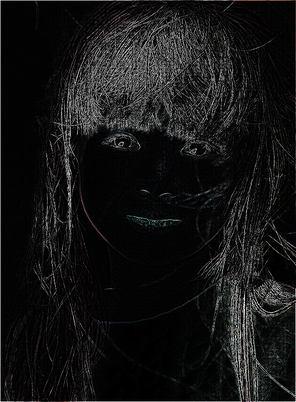} & \includegraphics[scale=0.524]{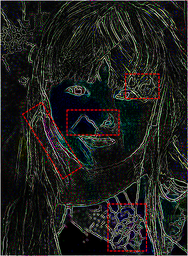} & \includegraphics[scale=0.194]{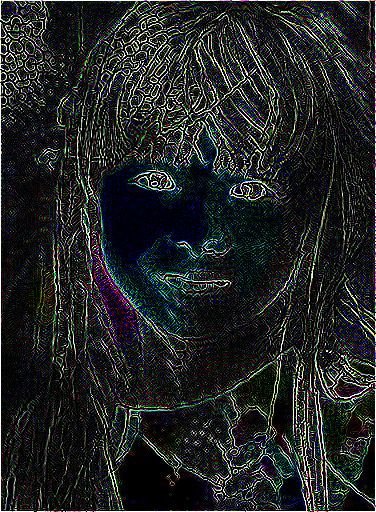}\tabularnewline
{\small{}(d) Content Laplacian} & {\small{}(e) Gatys-style Laplacian} & {\small{}(f) Lapstyle Laplacian}\tabularnewline
\end{tabular}\caption{\label{fig:lap-girl-cmp} Correspondences between Laplacian deviations
and image distortions. (a), (d): the content image and its Laplacian.
(b), (e): the stylized image by Gatys-style and its Laplacian. Major
Laplacian deviations in (e) are highlighted; the corresponding regions
in (b) are severely distorted. (c), (f): the stylized image by Lapstyle
and its Laplacian. Due to the Laplacian loss term, the major Laplacian
deviations are alleviated in (f), and the corresponding regions in
(c) become more faithful.}
\end{figure}

In the domain of computer vision, the Laplacian operator is widely
used to detect edges (regions of rapid intensity changes) \cite{cv-prince,imageproc}.
The Laplacian operator extracts a Laplacian matrix (a \emph{Laplacian}
in short) that contains the second-order variations in an image that
are prominent to human perception \cite{poisson,lap-patch}. These
second-order variations often correspond to detail structures. As
an example, Fig. \ref{fig:lap-girl-cmp}(d) is the Laplacian obtained
from the previous content image, which makes explicit important edges
and boundaries within the content image. Large deviations in the Laplacian
usually correspond to large distortions or artifacts in the stylized
image. For example, the highlighted regions in Fig. \ref{fig:lap-girl-cmp}(e)
correspond to distorted regions in Fig. \ref{fig:lap-girl-cmp}(b).
Naturally, by imposing a new constraint that requires the stylized
image to have a similar Laplacian to the content image, the stylized
image may better preserve the detail structures.

An image Laplacian constraint has the advantage that, since it is
applied to the second derivatives of image pixels, many images correspond
to the same Laplacian, and each of the image is a solution of this
Laplacian constraint. After putting other constraints back, an image
well satisfying all constraints may still exist. Poisson Image Editing
\cite{poisson} is an example of incorporating a Laplacian constrant
with some boundary constraints to transfer a region seamlessly. In
other words, a Laplacian constraint is flexible and compatible with
other constraints, e.g. the aforementioned content and style feature
constraints. Motivated by these virtues of the Laplacian, we propose
to add a \emph{Laplacian loss} into neural style transfer, to steer
the stylized image towards having a similar Laplacian to that of the
content image. The Laplacian loss is defined as the mean-squared distance
between the two Laplacians. Minimizing this loss drives the stylized
image to have similar detail structures as the content image. Meanwhile,
the stylized image will still be rendered in the new style. This enhanced
neural style transfer method is named \textbf{Lapstyle}. Fig. \ref{fig:lap-girl-cmp}(c)
show that in the presence of the Laplacian loss, severe Laplacian
deviations are eliminated or alleviated, and the stylized image is
more faithful to the original content image.

The Laplacian loss is computed by a small two-layer fixed CNN. This
network consists of an average pooling layer and a prespecified convolutional
layer, and can be easily plugged into almost all existing style transfer
methods to better preserve the detail structures of the content image.
The computational overhead of the Laplacian loss to an existing method
is negligible. Moreover, we have released our source code of Lapstayle\footnote{\url{https://github.com/askerlee/lapstyle}}
to facilitate further research.

We evaluated Lapstyle on various test images under different conditions.
Experimental results show that compared to the popular Gatys-style
method \cite{gatys16}, Lapstyle always produces stylized images more
faithful to the content images in low-level structures, with much
less artifacts and distortions, and almost equally ``stylish''.
In addition, we measured how the presence of the Laplacian loss term
impacts all types of losses in the optimization objective, and confirmed
that the Laplacian loss term drastically reduces the Laplacian loss
between the content and stylized images, at the cost of only slight
increases of the content and style losses. This supports the above
empirical observations quantitatively.

A natural extension to the Laplacian loss is to combine multiple losses
with pooling layers of different sizes, to capture different scales
of the image detail structures. We investigated several combinations
of two Laplacian losses, and presented an example improved by the
Laplacian combinations.

To sum up, the contributions of this work are threefold:
\begin{enumerate}[leftmargin=2em,topsep=2pt,itemsep=0.4ex]
\item We propose a novel detail-preserving loss named \emph{Laplacian loss}
for neural style transfer, which steers the synthesized image towards
having similar low-level structures as the content image, while being
flexible to allow the image to be rendered in the new style. Augmented
by the Laplacian loss, a new style transfer method named Lapstyle
is obtained. Extensive experiments show that Lapstyle generates stylized
images that are more appealing with less artifacts.
\item The Laplacian loss is computed through a small add-on CNN. This loss
can be easily incorporated into almost any existing style transfer
methods for better preserving the detail structures of the content
image.
\item Multiple Laplacian losses with different granularities can be combined
to capture the image detail structures at different scales. We have
empirically investigated several combinations and observed improvements
on some examples.
\end{enumerate}

\section{Related Work}

The task of style transfer has been studied in computer vision and
graphics communities for a long time, under the name of ``texture
transfer''. Some early works synthesize new images by sampling pixels
or image patches according to certain similarity metrics \cite{nonparam-samp,tree,graphcut,example}.
Another line of works first extract a set of feature maps from an
image by applying manually designed filters, and then transform the
feature maps of a synthesized image to match that of a style image
\cite{pyramid,wavelet}.

An important inspiration for the present work is Poisson Image Editing
\cite{poisson}, which aims to transfer a source region in an image
to a target region in another image seamlessly. Poisson Image Editing
restrains the target regions to have the same Laplacian as the source
region, and fixes the boundary pixels to be the original pixels around
the target region. As the Laplacian constraint has a big set of solutions,
given the additional boundary constraint, a solution still exists.
This solution is a transferred region fitted in the background in
the target image with smooth transitions. 

Agarwala et al. \cite{montage} and Darabi et al. \cite{melding}
incorporate gradient terms into the patch similarity metric for matching
structurally coherent patches and blending a region into another image
with higher consistency. Recently, Lee et al. \cite{lap-patch} propose
to complete missing regions in an image with a Laplacian pyramid,
i.e. stacked Laplacian filters in decreasing granularities. This method
selects patches that are structurally coherent by exploiting the property
that image Laplacians preserve the detail structures.

All the above methods are traditional style transfer methods, as they
either operate at the pixel level, or use manully designed filters.
Moreover, non-parametric, ad-hoc procedures are usually involved.
Hence they can hardly capture and preserve high-level semantic information,
and the manually designed components limit their generalization ability.
As a result, the synthesized images are often globally incongruous
or of low quality.

Gatys et al. \cite{gatys16} propose to use a convolutional neural
network (CNN) to extract features for style transfer, namely the Neural
Style Transfer method. Different layers of CNNs capture information
at different levels of abstraction. Constraints with respect to different
levels of features are integrated in a single CNN as different loss
functions. In particular, the style to be transferred is captured
as Gram correlation matrices of the style image features. In this
framework, image synthesis reduces to the optimization within the
CNN. This approach elegantly addresses the style transfer problem
with high-quality output images. Gatys et al. \cite{gatys17} make
two extensions to this method: 1) better control the spatial consistency
by using a guided Gram loss, and 2) better preserve the color distribution
of the content image, by incorporating an extra luminance channel
or matching colour histograms.

Johnson et al. \cite{faststyle} propose a CNN for transforming images,
named the transformation network. It is trained with the loss function
as in \cite{gatys16}, but the parameters to be trained are the CNN
weights, instead of an input image. For each style image, a dedicated
transformation network is trained. To get a stylized output image,
one only needs to feed the content image into this CNN and take a
forward pass, which is much faster than \cite{gatys16}. Li et al.
\cite{mrf} apply the traditional idea of patch-matching to CNN features.
They propose a Markov Random Field loss function that drives the CNN
feature patches in the synthesized image to approximate their best
matches in the style image. This method, namely MRF-CNN, is extended
by Champandard \cite{doodle}, in which user-provided segmentation
information is incorporated to improve the local consistency of stylized
images.

More recently, Luan et al. \cite{photo} propose a method that can
achieve photorealistic style transfer, as a post-processing step of
Neural Style Transfer. They first extract locally affine functions
using Matting Laplacian\footnote{This `Laplacian' refers to the graph laplacian, which is irrelevant
to the image Laplacian used in this work.} of Levin et al. \cite{matt}, then fine-tune the stylized image to
fit these affine functions. In the examples they presented, the stylized
images are photorealistic and well retain the detail structures of
the content image. On our test images (Section \ref{sec:Experiments}),
this method produces images that are much more consistent to the content
images, but their stylishness is greatly reduced.

\section{Method}

\subsection{Neural Style Transfer\label{subsec:Neural-Style-Transfer}}

Given a content image $\boldsymbol{x}_{c}$ and a style image $\boldsymbol{x}_{s}$,
let their corresponding CNN features at layer $l$ be denoted as $F_{l}(\boldsymbol{x}_{c})$
and $F_{l}(\boldsymbol{x}_{s})$, respectively. Suppose there are
$N_{l}$ filters in layer $l$, then $F_{l}(\boldsymbol{x})\in\mathcal{\Re}^{M_{l}(x)\times N_{l}}$
is a matrix with $N_{l}$ columns. Each column of $F_{l}(\boldsymbol{x})$
is a vectorized feature map, i.e. the response of a filter in layer
$l$, containing $M_{l}(\boldsymbol{x})=H_{l}(\boldsymbol{x})\cdot W_{l}(\boldsymbol{x})$
elements, where $H_{l}(\boldsymbol{x})$ and $W_{l}(\boldsymbol{x})$
are the height and width of each feature map, respectively, depending
on the size of the input image $\boldsymbol{x}$.

Neural style transfer generates an image $\hat{\boldsymbol{x}}$ that
depicts the content of image $\boldsymbol{x}_{c}$ in the style of
$\boldsymbol{x}_{s}$, by minimizing the following loss function of
$\hat{\boldsymbol{x}}$ \cite{gatys16,gatys17}:
\begin{equation}
\mathcal{L}_{\textrm{total}}=\alpha\mathcal{L}_{\textrm{content}}+\beta\mathcal{L}_{\textrm{style}},\label{eq:gatys-obj}
\end{equation}
where the content loss $\mathcal{L}_{\textrm{content}}$ is the mean-squared
distance between the feature maps of $\boldsymbol{x}_{c}$ and $\hat{\boldsymbol{x}}$
at a prespecified \emph{content layer} $l_{c}$:
\begin{equation}
\mathcal{L}_{\textrm{content}}=\frac{1}{N_{l_{c}}M_{l_{c}}(\boldsymbol{x}_{c})}\sum_{ij}(F_{l_{c}}(\hat{\boldsymbol{x}})-F_{l_{c}}(\boldsymbol{x}_{c}))_{ij}^{2},\label{eq:content-loss}
\end{equation}
and the style loss $\mathcal{L}_{\textrm{style}}$ measures the distributional
difference of the feature maps of $\boldsymbol{x}_{c}$ and $\hat{\boldsymbol{x}}$
at several prespecified \emph{style layers}:
\begin{align}
\mathcal{L}_{\textrm{style}} & =\sum_{l}w_{l}E_{l}(\hat{\boldsymbol{x}},\boldsymbol{x}_{s})\nonumber \\
E_{l}(\hat{\boldsymbol{x}},\boldsymbol{x}_{s}) & =\frac{1}{4N_{l}^{2}}\sum_{ij}(G_{l}(\hat{\boldsymbol{x}})-G_{l}(\boldsymbol{x}_{s}))_{ij}^{2},\label{eq:style-loss}
\end{align}
where $w_{l}$ is the weight of layer $l$, and $G_{l}(\boldsymbol{x})=\frac{1}{M_{l}(\boldsymbol{x})}F_{l}(\boldsymbol{x})^{\T}F_{l}(\boldsymbol{x})\in\Re^{N_{l}\times N_{l}}$
is the Gram matrix of the feature maps at layer $l$. The $i,j$-th
element of $G_{l}(\boldsymbol{x})$ measures the correlation between
the $i$-th and $j$-th feature maps, i.e. how often the $i$-th and
$j$-th filters are simultaneously activated across small image patches. 

The style loss on Gram matrices is essential to the impressive performance
of Neural style transfer. The intuitions behind this formulation are
as follows. A style or texture usually contains local and repetitive
combinations of multiple visual elements. Suppose different visual
elements activate different convolutional filters, then repetitive
local co-occurrence of these elements correspond to repetitive co-activation
patterns in the feature maps, and consequently, yield large positive
numbers in certain locations in the Gram matrix. By forcing the stylized
image $\hat{\boldsymbol{x}}$ to have similar Gram matrices to $\boldsymbol{x}_{s}$,
$\hat{\boldsymbol{x}}$ has to generate similar filter co-activation
patterns, which probably indicates that $\hat{\boldsymbol{x}}$ possesses
similar styles and textures. Different style layers are used to capture
the styles and textures at different granularities.

Neural style transfer methods usually employ a pretrained 19-layer
VGG network to extract features and perform the optimization, because
VGG preserves more information at the convolutional layers \cite{explore}.
In the original work \cite{gatys16}, Gatys et al. chose `conv4\_2'
as the content layer, and `conv1\_1\textquoteright , `conv2\_1\textquoteright ,
`conv3\_1\textquoteright , `conv4\_1\textquoteright{} and `conv5\_1\textquoteright{}
as the style layers.\footnote{In their implementation, they actually used `relux\_y' in place of
each `convx\_y', e.g. `relu\_4\_2' instead of `conv4\_2', where negative
responses are truncated to 0.}

The optimization (image synthesis) proceeds as follows:
\begin{enumerate}[leftmargin=2em,topsep=2pt,itemsep=0.6ex]
\item Set the input of the CNN as the content image $\boldsymbol{x}_{c}$.
Do a forward propagation and save the response $F_{l_{c}}(\boldsymbol{x}_{c})$
of the content layer;
\item Set the input of the CNN as the style image $\boldsymbol{x}_{s}$.
Do a forward propagation. Compute and save the Gram matrices $\{G_{l}(\boldsymbol{x}_{s})\}$
of all style layers;
\item Initialize $\hat{\boldsymbol{x}}$. $\hat{\boldsymbol{x}}$ could
be randomly initialized, or copied from the content image;
\item Iterate until reaching $N$ iterations:
\begin{enumerate}
\item Do a forward propagation, and compute the total loss;
\item Do a backward propagation to update $\hat{\boldsymbol{x}}$.
\end{enumerate}
\end{enumerate}

\subsection{Laplacian Operator}

The Laplacian operator $\triangle$ of a function $f$ is the sum
of all unmixed second partial derivatives:
\[
\triangle f=\sum_{i}\frac{\partial^{2}f}{\partial x_{i}^{2}}.
\]

The Laplacian filter \cite{cv-prince} is the discrete approximation
to the two dimensional Laplacian operator, given by 
\[
D=\left[\begin{array}{ccc}
0 & -1 & 0\\
-1 & 4 & -1\\
0 & -1 & 0
\end{array}\right].
\]
 The Laplacian matrix (``Laplacian'' in short) of an image $\boldsymbol{x}$
is obtained by convolving the image with $D$, denoted by $D(\boldsymbol{x})$.
At regions where adjacent pixel values change drastically, the convolution
will produce a response of high magnitude, regardless of the direction
of the change. In regions where the change is flat, the response is
zero \cite{imageproc}. Hence the Laplacian operator is widely used
for edge detection and extraction of detail structures in an image
\cite{lap-patch}.

\subsection{Laplacian Loss and Lapstyle Objective}

\begin{figure}

\begin{centering}
\hspace*{0em}\includegraphics[scale=0.4]{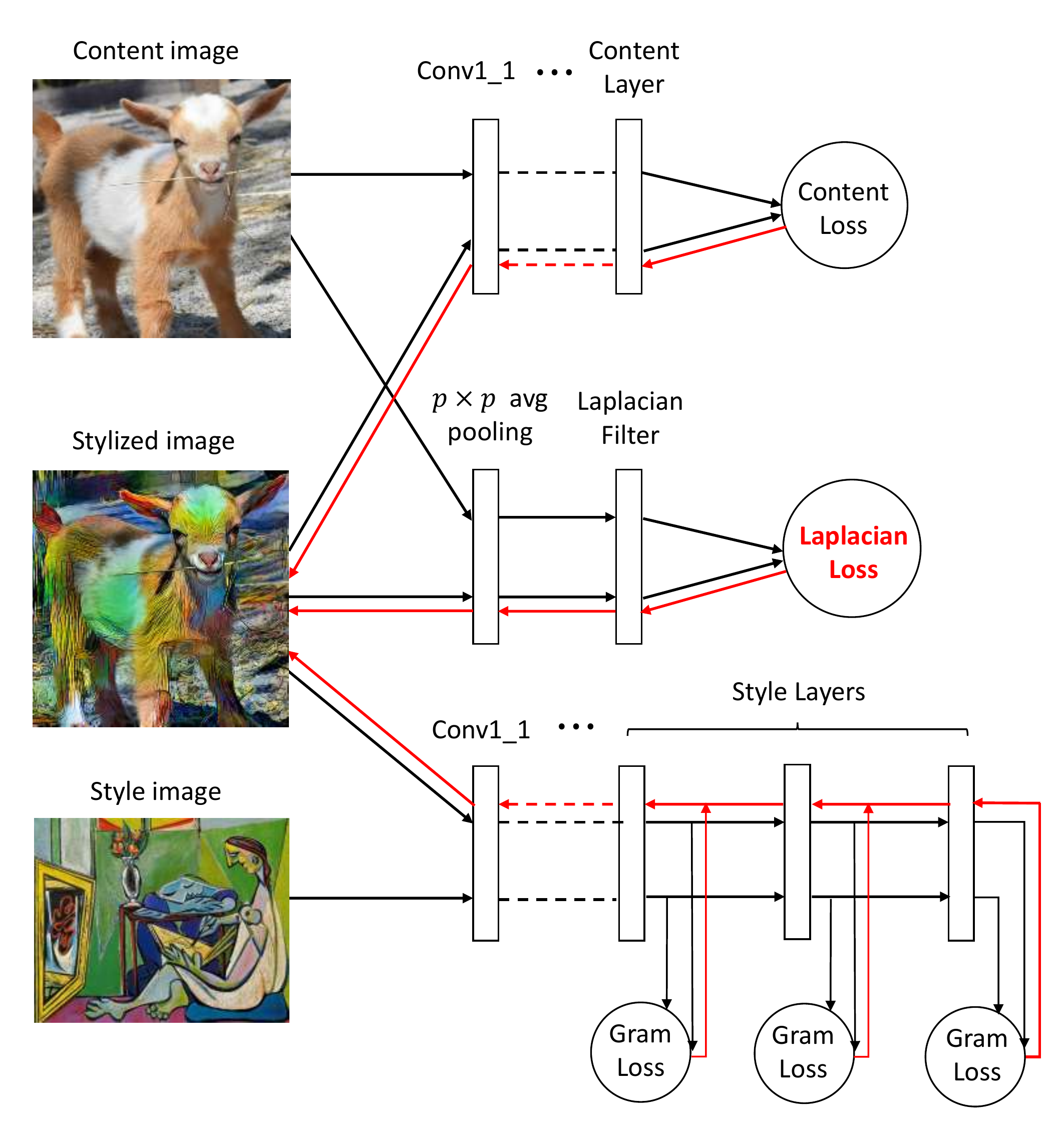}\caption{\label{fig:arch}Network Architecture of Lapstyle. Black and red lines
are forward and backward passes, respectively. Dashed lines indicate
that there are unshown intermediate layers.}
\par\end{centering}
\end{figure}

Given two images $\boldsymbol{x}_{c}$ and $\hat{\boldsymbol{x}}$,
we use a \emph{Laplacian loss} $\mathcal{L}_{\textrm{lap}}$ to measure
the difference between their Laplacians:
\begin{equation}
\mathcal{L}_{\textrm{lap}}=\sum_{ij}(D(\boldsymbol{x}_{c})-D(\hat{\boldsymbol{x}}))_{ij}^{2}.\label{eq:laploss}
\end{equation}

In order to force the stylized image to possess similar detail structures
as the content image, we augument the style transfer objective \eqref{eq:gatys-obj}
with the Laplacian loss $\mathcal{L}_{\textrm{lap}}$ of $\hat{\boldsymbol{x}}$
and $\boldsymbol{x}_{c}$:
\begin{equation}
\mathcal{L}_{\textrm{total}}=\alpha\mathcal{L}_{\textrm{content}}+\beta\mathcal{L}_{\textrm{style}}+\gamma\mathcal{L}_{\textrm{lap}},\label{eq:lapstyle-obj}
\end{equation}
where $\gamma$ is a tunable parameter controlling the strength of
the Laplacian loss. In situations where the output image is severly
distorted, we could increase $\gamma$ to demand a more faithful stylization.

The new optimization objective \eqref{eq:lapstyle-obj} and the corresponding
neural style transfer method is named \emph{Lapstyle}. For the Lapstyle
objective, the optimization procedure in Section \ref{subsec:Neural-Style-Transfer}
remains unchanged, except that in Step 1 the Laplacian $D(\boldsymbol{x}_{c})$
needs to be saved. Fig. \ref{fig:arch} presents the overall structure
of Lapstyle.

When computing the Laplacian loss, two practical issues are addressed
as follows:

\textbf{Laplacians on different channels}\quad{}The input image $\boldsymbol{x}$
has three RGB channels. Ideally, to detect edges, each channel should
be considered separately, i.e. each channel convolves with the Laplacian
filter, yielding three intermediate Laplacians. A value with a large
magnitude in any of the three Laplacians suggests an edge, regardless
of its sign. To this end, the final Laplacian $D(\boldsymbol{x})$
used in the Laplacian loss should be the sum of the absolute values
of the three Laplacians \cite{imageseg}: $D(\boldsymbol{x})=\left\Vert D(\boldsymbol{x}^{R})\right\Vert +\left\Vert D(\boldsymbol{x}^{G})\right\Vert +\left\Vert D(\boldsymbol{x}^{B})\right\Vert $.
However, this formulation involves some technical difficulties in
implementation. For simplification, we adopt an approximation using
their sum as the final Laplacian:
\begin{equation}
D(\boldsymbol{x})=D(\boldsymbol{x}^{R})+D(\boldsymbol{x}^{G})+D(\boldsymbol{x}^{B}).\label{eq:lapsum}
\end{equation}

This summation is automatically calculated by convolving the three
channels altogether with a Laplacian convolutional filter. We have
not observed noticable degradation of image quality with this approximation.

\textbf{Smoothing with a pooling layer}\quad{}The Laplacian filter
is sensitive to small perturbations in the input image, and smoothing
the input image can make the Laplacian loss better reflect its true
detail structures. Hence we add a $p\times p$ average pooling layer
before the Laplacian layer for smoothing. Pooling also reduces the
memory overhead of the Laplacian loss to $1/p^{2}$ of that on the
original image.

\subsection{Combining Multiple Laplacians}

When the pooling layer has a wider kernel, it condenses a larger area
into one pixel. A Laplacian on top of such a pooling layer may capture
structures in larger regions. Hence, combining multiple Laplacian
losses over increasingly dilated pooling layers may capture detail
structures in different granularities. The optimization objective
is accordingly extended to
\begin{equation}
\mathcal{L}_{\textrm{total}}=\alpha\mathcal{L}_{\textrm{content}}+\beta\mathcal{L}_{\textrm{style}}+\sum_{k}\gamma_{k}\mathcal{L}_{\textrm{lap}k},\label{eq:lapstyle-obj-1}
\end{equation}
where $\mathcal{L}_{\textrm{lap}k}$ is the Laplacian loss on the
images pooled by an average pooling layer of size $p_{k}\times p_{k}$,
and $\gamma_{k}$ is its weight.

\begin{figure*}
\begin{centering}
\begin{tabular}{cccc}
\includegraphics[scale=0.12]{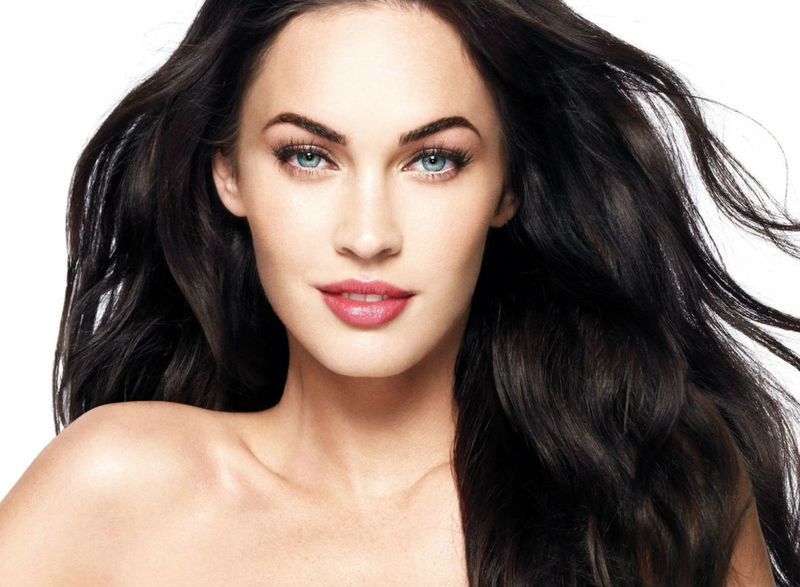} & \includegraphics[scale=0.24]{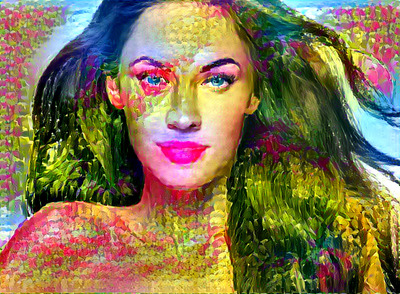} & \includegraphics[scale=0.24]{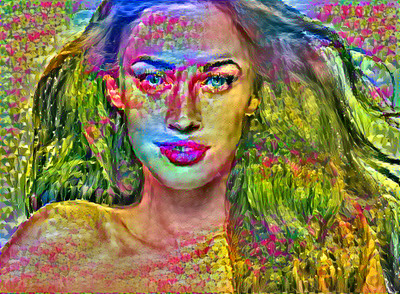} & \includegraphics[scale=0.21]{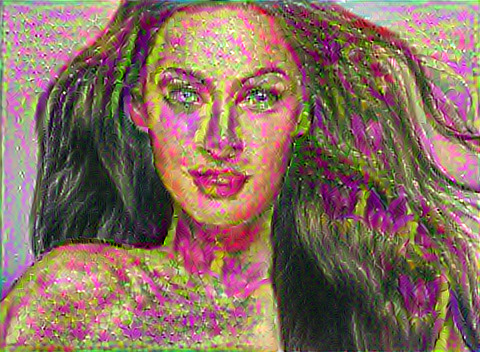}\tabularnewline
Content image & Gatys-style using L-BFGS & Gatys-style using Adam & MRF-CNN\tabularnewline[1em]
\includegraphics[scale=0.92]{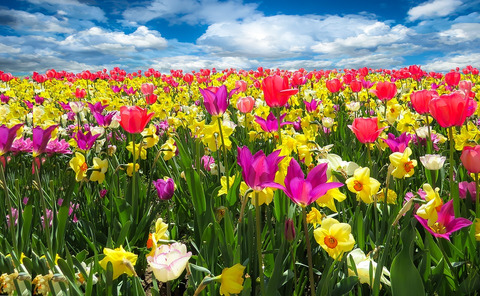} & \includegraphics[scale=0.24]{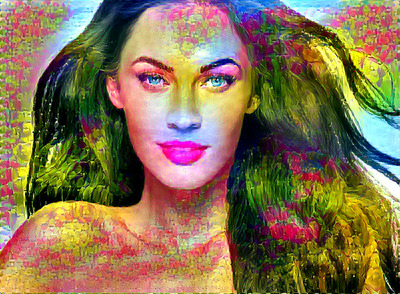} & \includegraphics[scale=0.24]{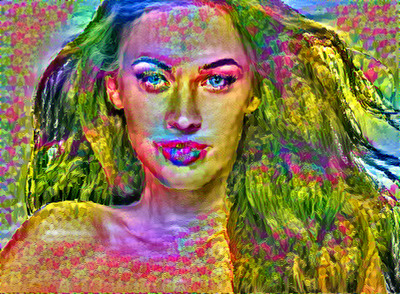} & \includegraphics[scale=0.21]{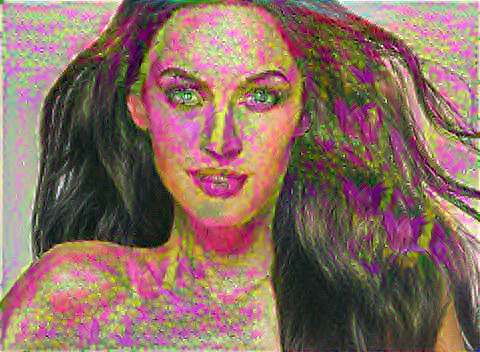}\tabularnewline
Style image & \multicolumn{3}{c}{\raisebox{2ex}{$\underbrace{\hspace{20em}}_{\textnormal{\normalsize Corresponding Lapstyle extensions}}$}}\tabularnewline
\end{tabular}
\par\end{centering}
\caption{\label{fig:implements}Style transfer results using differnt implementations
(first row) and their corresponding Lapstyle extensions (second row).
L-BFGS implementation of Gatys-style produces the best stylized image
among the original implementations. Lapstyle extensions always produce
better images than the original ones.}
\end{figure*}

\subsection{Implementation Details}

The Laplacian filter is a 3x3 convolutional layer with fixed weights.
Hence the Laplacian loss can be conveniently implemented in any mainstream
deep learning frameworks and incorporated into most existing neural
style transfer methods. 

We have made extensions to three commonly used neural style transfer
packages, and found that Gatys-style implemented using the L-BFGS
optimization method yields the best stylized images. See Section \ref{sec:Experiments}
for more details.

\section{Experiments}

In this section, we show the style transfer results on various content
and style images, using the Lapstyle extensions to three commonly
used neural style transfer packages. Compared to the the original
packages, Lapstyle constantly yields more appealing images that better
preserve the detail structures of the content image.

\label{sec:Experiments}

\textbf{Baselines.}\quad{}Four commonly used neural style transfer
packages are used as baseline methods. Among them, two are implementations
of Gatys-style \cite{gatys16}, one implemented by Justin Johnson\footnote{\url{https://github.com/jcjohnson/neural-style}},
using the L-BFGS optimization method, and the other by Anish Athalye\footnote{\url{https://github.com/anishathalye/neural-style}},
using the Adam adaptive optimization method \cite{adam}. The third
package is MRF-CNN \cite{mrf} implemented by Alex J. Champandard\footnote{\url{https://github.com/alexjc/neural-doodle}}.
The fourth package is the recent Photorealistic Style Transfer \cite{photo}\footnote{\url{https://github.com/luanfujun/deep-photo-styletransfer}},
referred to as Photo-style in the following.

\textbf{Lapstyle Settings.}\quad{}Each of these three packages were
extended with the Laplacian loss to get their Lapstyle counterparts.
Due to the distinctiveness of each implementation, different sizes
of the pooling layer and the Laplacian weight were adopted for different
baselines. For the extension to the L-BFGS implementation of Gatys-style,
we use a pooling size $p=4$, the Laplacian weight $\gamma=100$,
and the total iterations $N=1000$. For the extension to the Adam
implementation of Gatys-style, we fix $p=2,\gamma=200,N=1000$. For
the extension to MRF-CNN, we set $p=2,\gamma=100,N=300$. For Photo-style,
we set $\lambda=1000$.

Fig. \ref{fig:implements} compares the stylized images using a same
test pair. Clearly, the L-BFGS implementation of Gatys-style and its
corresponding Lapstyle extension yields the best images. In fact,
the L-BFGS implementation of Gatys-style has been observed to consistently
perform the best on a series of content and style images (the comparisons
are not shown in this paper due to space limitations). Hence in the
following, we take the L-BFGS implementation of Gatys-style as the
main baseline against Lapstyle, and simply denote it as Gatys-style.

\subsection{Qualititive Results}

\begin{figure*}[p]
\hspace*{-0.5cm}%
\begin{tabular}{cccc}
\includegraphics[scale=0.15]{images/megan}\hspace{-4.1em}\includegraphics[scale=0.15]{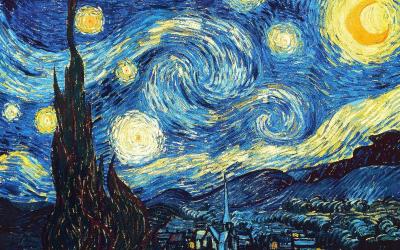} & \includegraphics[scale=0.24]{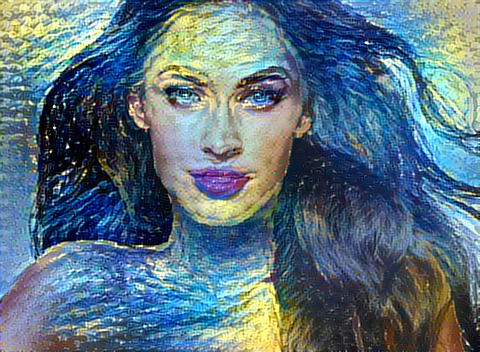} & \includegraphics[scale=0.24]{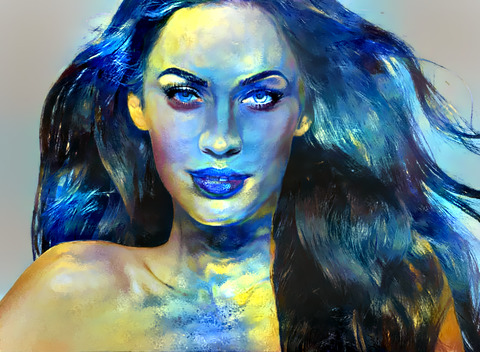} & \includegraphics[scale=0.24]{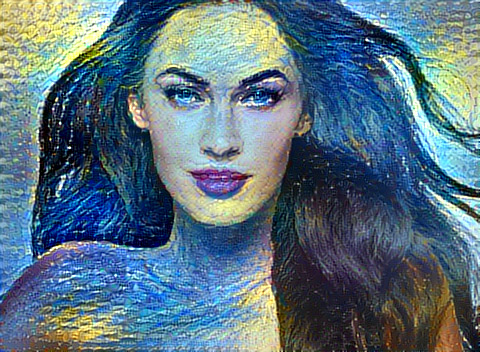}\tabularnewline
\noalign{\vskip0.5cm}
\includegraphics[scale=0.78]{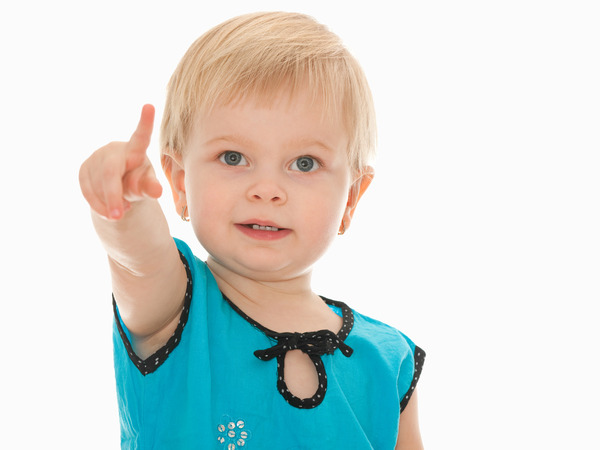}\hspace{-3em}\includegraphics[scale=0.1]{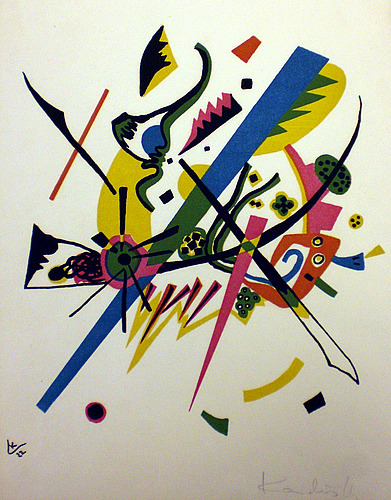} & \includegraphics[scale=0.22]{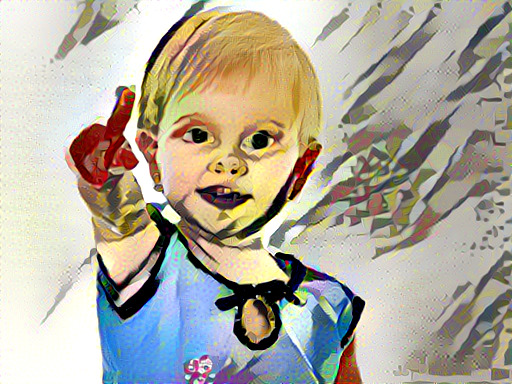} & \includegraphics[scale=0.22]{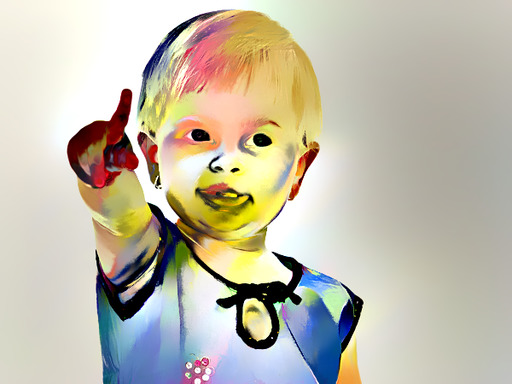} & \includegraphics[scale=0.22]{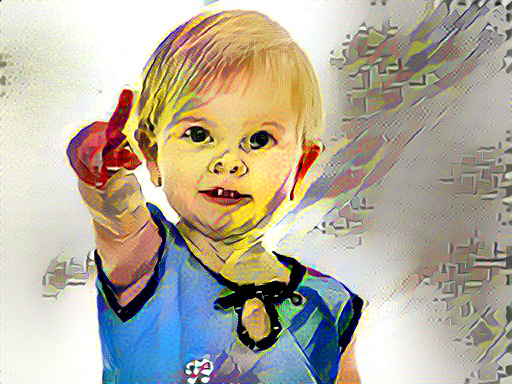}\tabularnewline
\noalign{\vskip0.5cm}
\includegraphics[scale=0.28]{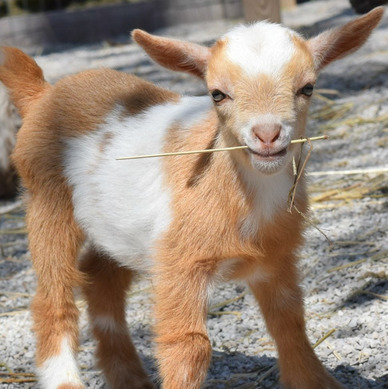}\hspace{-3em}\includegraphics[scale=0.2]{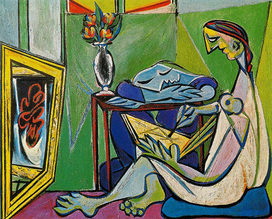} & \includegraphics[scale=0.28]{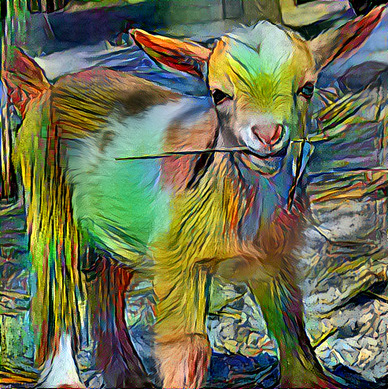} & \includegraphics[scale=0.28]{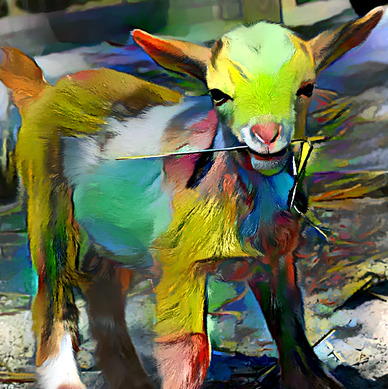} & \includegraphics[scale=0.28]{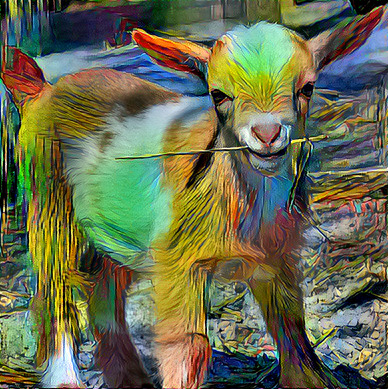}\tabularnewline
\noalign{\vskip0.5cm}
\includegraphics[scale=0.35]{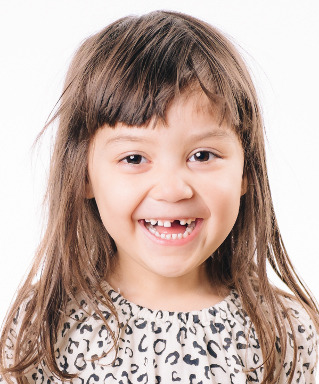}\hspace{-3.5em}\includegraphics[scale=0.15]{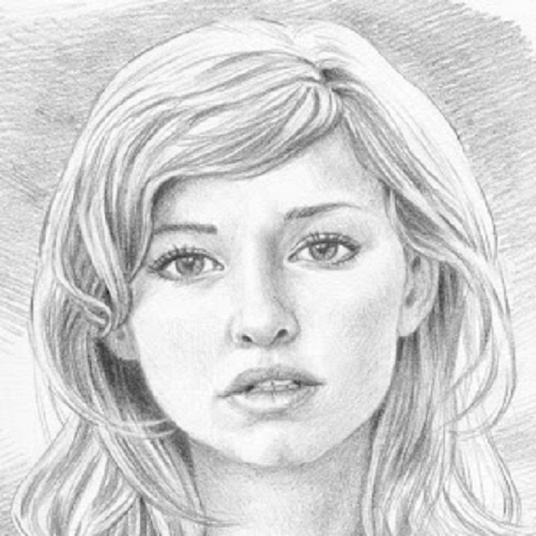} & \includegraphics[scale=0.35]{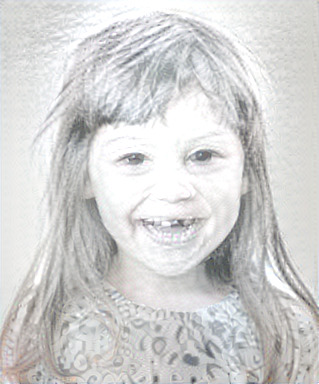} & \includegraphics[scale=0.35]{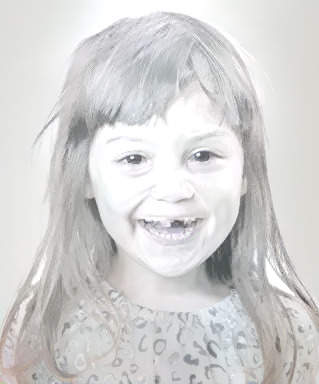} & \includegraphics[scale=0.35]{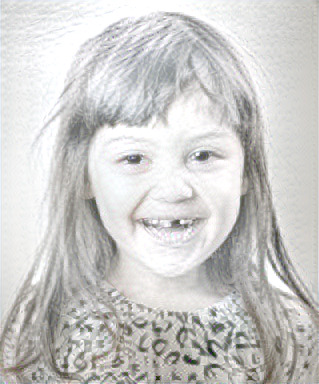}\tabularnewline
\noalign{\vskip0.5cm}
\includegraphics[scale=2.4]{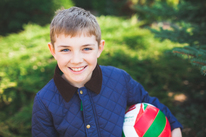}\hspace{-2.5em}\includegraphics[scale=0.1]{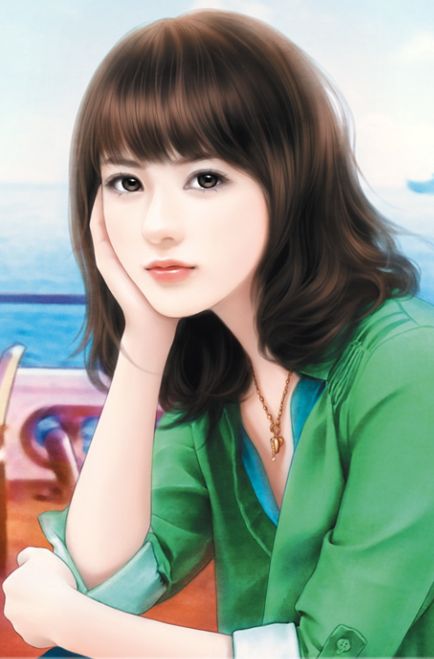} & \includegraphics[scale=0.15]{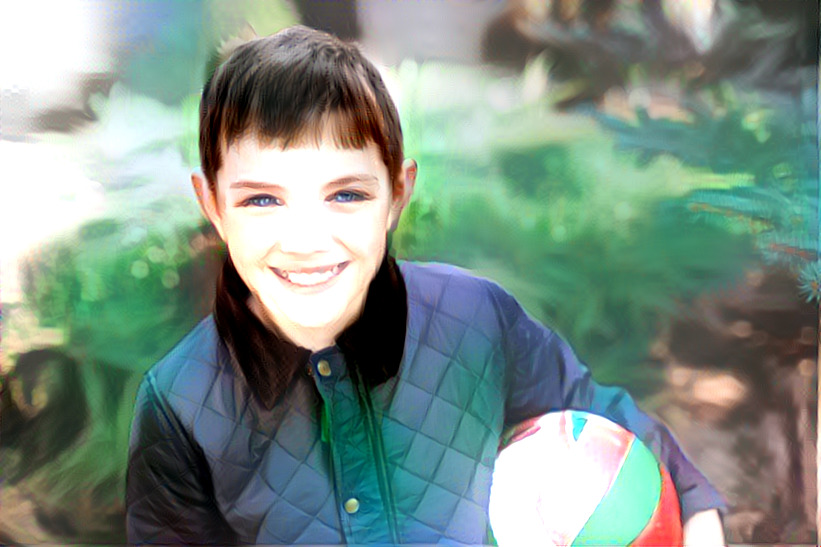} & \includegraphics[scale=0.15]{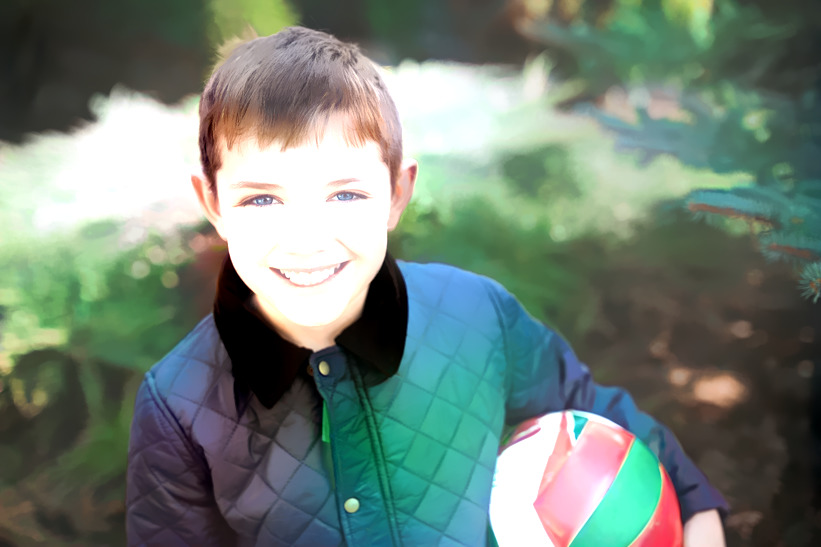} & \includegraphics[scale=0.15]{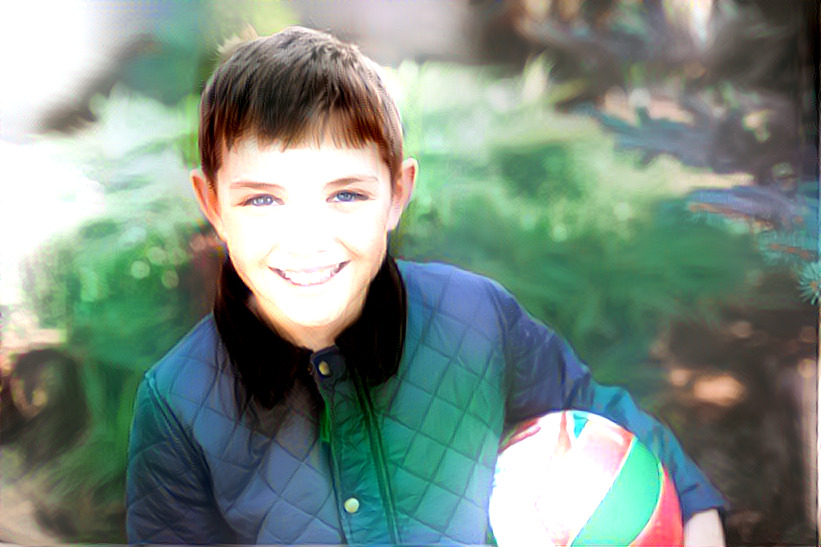}\tabularnewline
Content Image & Gatys-style & Photo-style & Lapstyle\tabularnewline
\end{tabular}

\caption{\label{fig:more-comp}Comparisons of Gatys-style, Photo-style and
Lapstyle. Compared to Gatys-style, Lapstyle always yields stylized
images with finer and better-preserved local details, and less artifacts
introduced from the style images. Photo-style yields images with least
artifacts among the three methods, but the finer details are often
lost.}
\end{figure*}

In Fig. \ref{fig:more-comp}, we qualitatively compare Lapstyle with
Gatys-style and Photo-style on a variety of style and content images\footnote{Some images are from \cite{faststyle,mrf}.}.

It can be observed that images synthesized by Lapstyle consistently
improve on those by Gatys-style. In particular, more delicate details,
e.g. contours and gradual color transitions are preserved, and much
less artifacts are introduced from the style images. In the meantime,
with the reinforced content details, the ``stylishness'' of these
images do not degrade noticeably. This shows that the Laplacian constraint
is a content constraint flexible enough to entertain style constraints.

The synthesized images by Photo-style have both advantages and disadvantages
compared to Lapstyle. On the one hand, the synthesized images are
smoother with less artifacts, for example the random patterns in the
background now disappear. On the other hand, the synthesized images
are less stylized and less natural, and the style transfer mostly
happened to the colors. These results suggest that the Matting Laplacian
is a strong content constraint which may expel the effects of the
style constraints on local structures\footnote{Reducing the weight $\lambda$ of the Matting Laplacian constrant
does not substantially improve this problem.}.
\begin{figure}[H]
\noindent \begin{centering}
\hspace*{-0.2cm}%
\begin{tabular}{ccc}
\includegraphics[scale=0.18]{images/megan_spring20} & \includegraphics[scale=0.15]{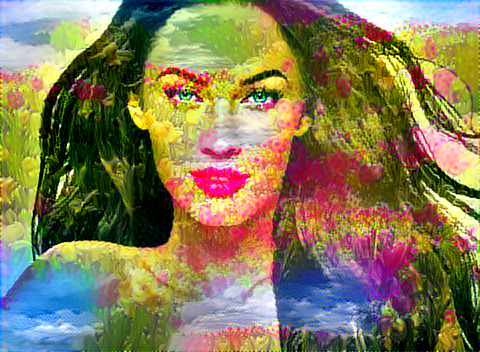} & \includegraphics[scale=0.15]{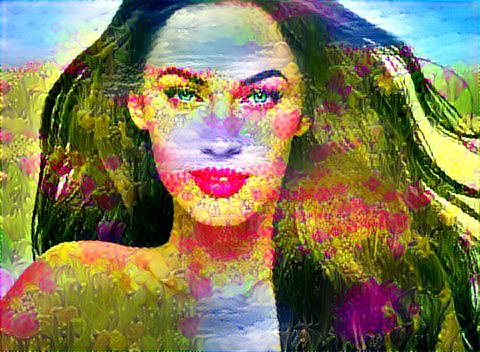}\tabularnewline
\includegraphics[scale=0.18]{images/goat_muse20_1_0-small} & \includegraphics[scale=0.18]{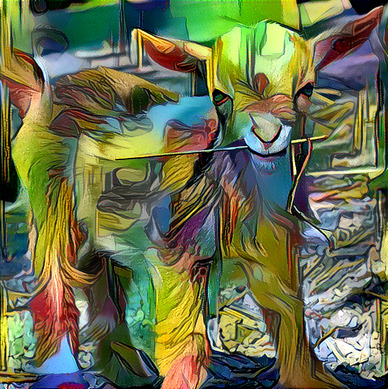} & \includegraphics[scale=0.18]{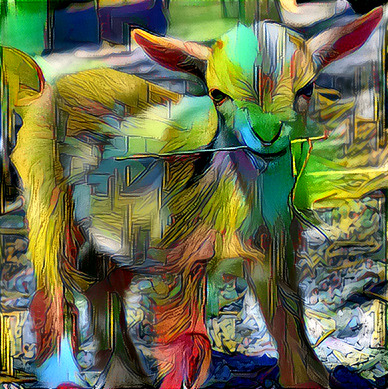}\tabularnewline
{\small{}(a) `conv4\_2'} & {\small{}(b) `conv2\_2'} & {\small{}(c) `conv2\_2' + Lapstyle}\tabularnewline
\end{tabular}
\par\end{centering}
\caption{\label{fig:conv22} Comparison of (a) Gatys-style with content layer
= `conv4\_2', (b) Gatys-style with content layer = `conv2\_2', and
(c) the Lapstyle extension of (b).}
\end{figure}

\textbf{Lower-level content layer.}\quad{}One may argue that content
details could be captured by using lower-level CNN filter banks as
the content layer \cite{gatys16} than the default layer `conv4\_2',
to achieve similar effects as does Lapstyle. Hence we experimented
to use `conv2\_2' as the content layer. Fig. \ref{fig:conv22} compares
the output images of using `conv4\_2', `conv2\_2' and `conv2\_2' +
Lapstyle, respectively, given two test pairs appeared before. When
`conv2\_2' is used as the content layer, the stylized images are apparently
of lower quality than using `conv4\_2'. Nevertheless, incorporating
the Laplacian loss still improves the stylized images in this setting.
This verifies that the Laplacian filter, being a single manually chosen
CNN filter, captures content details that can not be captured by the
pretrained low-level CNN filter banks.

\begin{figure}
\centering{}\hspace*{-0.25cm}%
\begin{tabular}{ccc}
\includegraphics[scale=0.2]{images/girlmrf} & \includegraphics[scale=1.7]{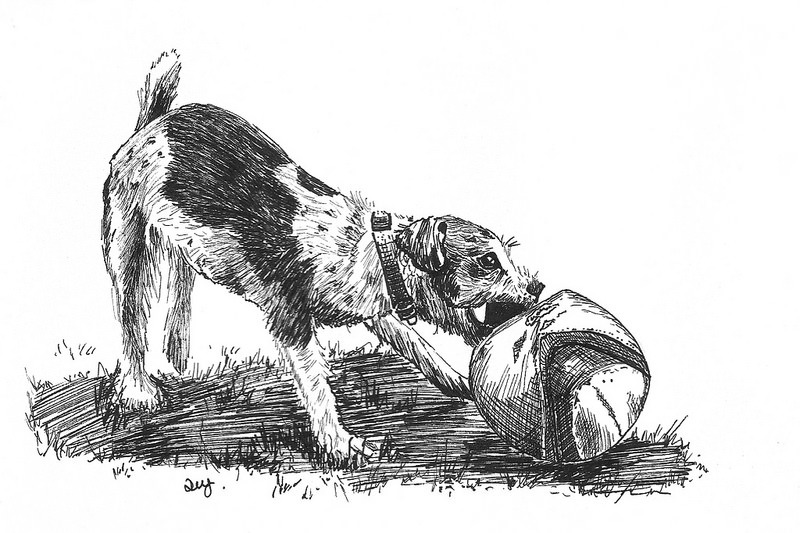} & \includegraphics[scale=0.2]{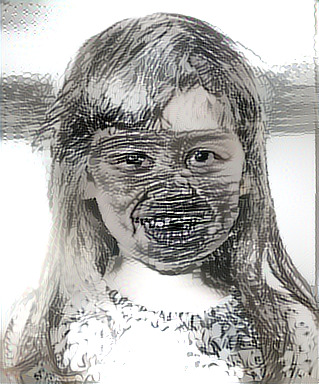}\tabularnewline
Content image & Style image & Gatys-style\tabularnewline
\includegraphics[scale=0.2]{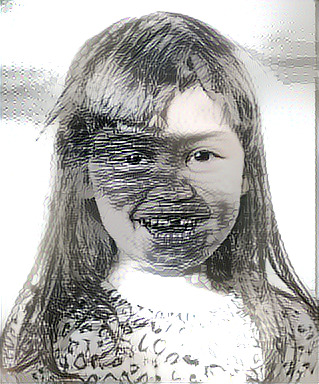} & \includegraphics[scale=0.2]{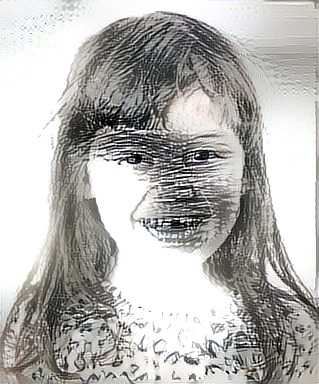} & \includegraphics[scale=0.2]{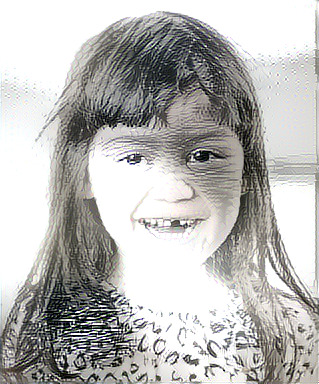}\tabularnewline
$p=2$ & $p=4$ (default) & $p_{1}=2,p_{2}=4$\tabularnewline
\includegraphics[scale=0.2]{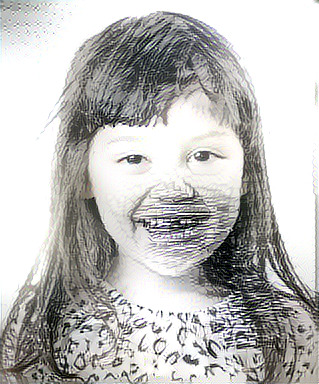} & \includegraphics[scale=0.2]{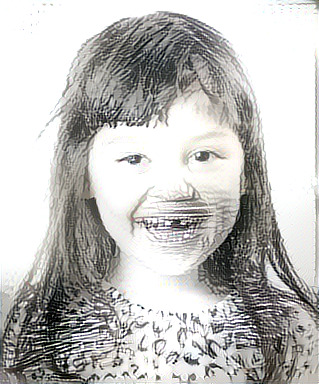} & \includegraphics[scale=0.2]{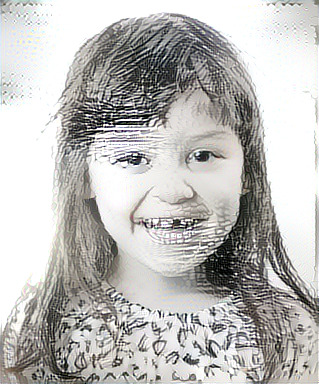}\tabularnewline
$p_{1}=2,p_{2}=8$ & $p_{1}=4,p_{2}=8$ & $p_{1}=4,p_{2}=16$\tabularnewline
\end{tabular}\caption{\label{fig:multi-lap}Results using one or two Laplacian losses of
different pooling sizes. The captions of the 2nd and 3rd rows give
the pooling layer sizes, e.g. $p_{1}=2,p_{2}=4$ means it uses a 2x2
pooled Laplacian loss combined with a 4x4 pooled Laplacian loss.}
\end{figure}

\textbf{Combining Multiple Laplacians.}\quad{}We experimented to
combine two Laplacian losses with different pooling sizes. Compared
with a single Laplacian loss, it led to improvement on some test images.
Fig. \ref{fig:multi-lap} presents such an example, where the best
stylization is achieved by combining a 4x4 pooled Laplacian loss with
a 16x16 pooled Laplacian loss. The default setting of a single 4x4
pooled Laplacian loss failed to remove some artifacts on the girl's
face. 

However, the actual effects of combined Laplacians often depend on
the particular content and style images. For fairness of comparison,
we did not seek the best combinations for each test pair, instead
in other experiments we simply used the default single 4x4 pooled
Laplacian loss, which brings about empirically stable improvements.

\begin{figure}
\centering{}%
\begin{tabular}{cc>{\centering}p{3cm}}
\includegraphics[scale=0.18]{images/girl2_cartoon2_20_0} & \includegraphics[scale=0.18]{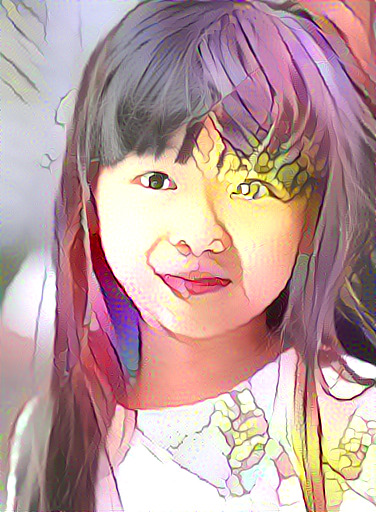} & \includegraphics[scale=0.23]{images/girl2_cartoon2_20_100}\tabularnewline
Gatys-style & Lapstyle with LoG & Pooling + Laplacian\tabularnewline
\end{tabular}\caption{\label{fig:log}Comparison of the extensions with the Laplacian of
Gaussian (LoG) filter and with pooling + Laplacian filter.}
\end{figure}

\textbf{Laplacian of Gaussian.}\quad{}The Laplacian of Gaussian filter
(LoG, i.e., Gaussian smoothed Laplacian filter) \cite{imageproc}
is commonly used in place of the vanilla Laplacian filter, as it is
more robust to noisy pixel variations. We tested to replace the pooling
+ Laplacian structure with the commonly used LoG of size 5x5. As seen
in Fig. \ref{fig:log}, it led to much weaker effects of preserving
detail structure and removing artifacts, compared to using an average
pooling layer. 

\subsection{Quantitative Analysis}

In order to quantitatively investigate how the Laplacian loss term
impacts the Laplacian, content and style losses of the synthesized
image, we computed the values of these losses with and without the
Laplacian loss, respectively, averaged over the five pairs of input
images in Fig. \ref{fig:more-comp}. The parameter settings were the
same as described before. All the losses are normalized by the initial
Laplacian loss. Table \ref{tab:Loss-changes} lists the mean Laplacian,
content, style and total losses in the first iteration and the last
iteration of the optimization, respectively. It then compares the
losses of the final images optimized with and without the Laplacian
loss term.

\begin{table}
\centering{}%
\begin{tabular}{>{\raggedright}p{1.15cm}|>{\centering}p{1.8cm}|>{\centering}m{0.8cm}|>{\centering}m{1cm}|>{\centering}m{0.8cm}|>{\centering}m{0.8cm}}
\hline 
\multicolumn{2}{c|}{} & {\footnotesize{}Total loss} & {\footnotesize{}Laplacian loss} & {\footnotesize{}Content loss} & {\footnotesize{}Style loss}\tabularnewline
\hline 
\multicolumn{2}{c|}{{\footnotesize{}Init (Iter-0) losses}} & {\footnotesize{}222.1 (185.2) } & {\footnotesize{}1.00 (0.00)} & {\footnotesize{}7.0 (3.3)} & {\footnotesize{}214.1 (184.1) }\tabularnewline
\hline 
\multirow{3}{1.15cm}{{\footnotesize{}Lapstyle at Iter-1000}} & {\footnotesize{}Losses} & {\footnotesize{}7.44 (5.58)} & {\footnotesize{}0.22 (0.33)} & {\footnotesize{}5.86 (4.44)} & {\footnotesize{}1.36 (0.93)}\tabularnewline
\cline{2-6} 
 & {\footnotesize{}Frac of total loss} & / & {\footnotesize{}2.96\%} & {\footnotesize{}78.8\%} & {\footnotesize{}18.3\%}\tabularnewline
\cline{2-6} 
 & {\footnotesize{}Ratio to Init losses} & {\footnotesize{}3.3\%} & {\footnotesize{}22\%} & {\footnotesize{}84\%} & {\footnotesize{}0.64\%}\tabularnewline
\hline 
\multirow{3}{1.15cm}{{\footnotesize{}Gatys-style at Iter-1000}} & {\footnotesize{}Losses} & {\footnotesize{}11.41 (8.64)} & {\footnotesize{}4.49 (3.74)} & {\footnotesize{}5.73 (4.30)} & {\footnotesize{}1.19 (0.73)}\tabularnewline
\cline{2-6} 
 & {\footnotesize{}Frac of total loss} & / & {\footnotesize{}39.4\%} & {\footnotesize{}50.2\%} & {\footnotesize{}10.4\%}\tabularnewline
\cline{2-6} 
 & {\footnotesize{}Ratio to Init losses} & {\footnotesize{}5.1\%} & {\footnotesize{}449\%} & {\footnotesize{}82\%} & {\footnotesize{}0.56\%}\tabularnewline
\hline 
\multicolumn{2}{>{\centering}m{3cm}|}{{\footnotesize{}Ratios between Lapstyle and Gatys-style losses at
Iter-1000}} & {\footnotesize{}1.53} & \textbf{\footnotesize{}0.049} & \textbf{\footnotesize{}1.02} & \textbf{\footnotesize{}1.14}\tabularnewline
\hline 
\end{tabular}\bigskip{}
\caption{\label{tab:Loss-changes}The losses and relative changes with and
without the Laplacian loss term in the optimization objective. The
standard deviation of each loss is shown in parentheses. All the losses
are normalized by the initial Laplacian loss.}
\end{table}

It can be seen from Table \ref{tab:Loss-changes} that, with the Laplacian
loss term, when the optimization finishes, the Laplacian loss was
\emph{reduced} to 22\% of its initial value, amounting to only 2.96\%
of the total loss; as a trade-off, the content and style losses were
2\% and 14\% more than those of Gatys-style, respectively. Without
this loss term, however, the final Laplacian loss \emph{increased}
to 449\% of its initial value, amounting to 39.4\% of the total loss,
which may indirectly reflect the various artifacts in the stylized
images. 

The above quantitative analysis shows that by incorporating the Laplacian
loss term, the Laplacian loss of the synthesized images is reduced
to a large extent, at the cost of slight increases of the content
and style losses. Thereby it can be concluded that 1) the image stylized
by Lapstyle is almost equally ``stylish'' as that produced by the
traditional Gatys-style method; 2) the content loss and the Laplacian
loss govern independent aspects of the content of the stylized image.

\section{Conclusions}

Neural Style Transfer \cite{gatys16} adopts an optimization objective
that is concerned to preserve only high-level CNN features of the
content image. Due to the abstractiveness of these features, the synthesized
images suffer from various artifacts and distortions. In this paper
we have presented a method named Lapstyle that reduces artifacts and
distortions, by incorporating a novel Laplacian loss term into the
optimization objective. This loss encourages the stylized image to
have a similar image Laplacian as that of the content image. The image
Laplacian captures the detail structures (edges, contours, gradual
color transitions, etc.) of the content image, and thus guides the
optimization towards a synthesized image that more faithfully retains
these details. We have empircally validated that images synthesized
by Lapstyle look more natural and appealing, and contain much less
artifacts than the traditional methods, while remaining almost equally
``stylish''. In addition, we have observed quantitatively that while
the Laplacian loss is drastically reduced, the content and style losses
only increase slightly. This confirms that the Laplacian loss term
does not disrupt the image stylization.

In future work, we would like to explore ways to incorporate the Laplacian
loss into Generative Adversarial Nets (GANs) \cite{gan,cyclegan},
to improve the quality of images generated by GANs.

\section*{Acknowlegement}

This research was funded by the National Research Foundation, Prime
Minister\textquoteright s Office, Singapore under its IRC@SG Funding
Initiative.\nocite{zhang2017relation,zhang2016learning,Nie:2012SIGMM,Nie:2011SIGIR}

\bibliographystyle{ACM-Reference-Format}
\bibliography{laplacian-style}


\begin{thebibliography}{00}


\ifx \showCODEN    \undefined \def \showCODEN     #1{\unskip}     \fi
\ifx \showDOI      \undefined \def \showDOI       #1{#1}\fi
\ifx \showISBNx    \undefined \def \showISBNx     #1{\unskip}     \fi
\ifx \showISBNxiii \undefined \def \showISBNxiii  #1{\unskip}     \fi
\ifx \showISSN     \undefined \def \showISSN      #1{\unskip}     \fi
\ifx \showLCCN     \undefined \def \showLCCN      #1{\unskip}     \fi
\ifx \shownote     \undefined \def \shownote      #1{#1}          \fi
\ifx \showarticletitle \undefined \def \showarticletitle #1{#1}   \fi
\ifx \showURL      \undefined \def \showURL       {\relax}        \fi
\providecommand\bibfield[2]{#2}
\providecommand\bibinfo[2]{#2}
\providecommand\natexlab[1]{#1}
\providecommand\showeprint[2][]{arXiv:#2}

\bibitem[\protect\citeauthoryear{Acharya and Ray}{Acharya and Ray}{2005}]%
        {imageproc}
\bibfield{author}{\bibinfo{person}{Tinku Acharya} {and} \bibinfo{person}{Ajoy~K
  Ray}.} \bibinfo{year}{2005}\natexlab{}.
\newblock \bibinfo{booktitle}{{\em Image processing: principles and
  applications}}.
\newblock \bibinfo{publisher}{John Wiley \& Sons}.
\newblock


\bibitem[\protect\citeauthoryear{Agarwala, Dontcheva, Agrawala, Drucker,
  Colburn, Curless, Salesin, and Cohen}{Agarwala et~al\mbox{.}}{2004}]%
        {montage}
\bibfield{author}{\bibinfo{person}{Aseem Agarwala}, \bibinfo{person}{Mira
  Dontcheva}, \bibinfo{person}{Maneesh Agrawala}, \bibinfo{person}{Steven
  Drucker}, \bibinfo{person}{Alex Colburn}, \bibinfo{person}{Brian Curless},
  \bibinfo{person}{David Salesin}, {and} \bibinfo{person}{Michael Cohen}.}
  \bibinfo{year}{2004}\natexlab{}.
\newblock \showarticletitle{Interactive digital photomontage}. In
  \bibinfo{booktitle}{{\em ACM Transactions on Graphics (ToG)}},
  Vol.~\bibinfo{volume}{23}. ACM, \bibinfo{pages}{294--302}.
\newblock


\bibitem[\protect\citeauthoryear{{Champandard}}{{Champandard}}{2016}]%
        {doodle}
\bibfield{author}{\bibinfo{person}{A.~J. {Champandard}}.}
  \bibinfo{year}{2016}\natexlab{}.
\newblock \showarticletitle{{Semantic Style Transfer and Turning Two-Bit
  Doodles into Fine Artworks}}.
\newblock \bibinfo{journal}{{\em ArXiv e-prints\/}} (\bibinfo{date}{March}
  \bibinfo{year}{2016}).
\newblock
\showeprint[arxiv]{cs.CV/1603.01768}


\bibitem[\protect\citeauthoryear{Darabi, Shechtman, Barnes, Goldman, and
  Sen}{Darabi et~al\mbox{.}}{2012}]%
        {melding}
\bibfield{author}{\bibinfo{person}{Soheil Darabi}, \bibinfo{person}{Eli
  Shechtman}, \bibinfo{person}{Connelly Barnes}, \bibinfo{person}{Dan~B
  Goldman}, {and} \bibinfo{person}{Pradeep Sen}.}
  \bibinfo{year}{2012}\natexlab{}.
\newblock \showarticletitle{Image melding: Combining inconsistent images using
  patch-based synthesis.}
\newblock \bibinfo{journal}{{\em ACM Transactions on Graphics (ToG)\/}}
  \bibinfo{volume}{31}, \bibinfo{number}{4} (\bibinfo{year}{2012}),
  \bibinfo{pages}{82--1}.
\newblock


\bibitem[\protect\citeauthoryear{Efros and Leung}{Efros and Leung}{1999}]%
        {nonparam-samp}
\bibfield{author}{\bibinfo{person}{Alexei~A. Efros} {and}
  \bibinfo{person}{Thomas~K. Leung}.} \bibinfo{year}{1999}\natexlab{}.
\newblock \showarticletitle{Texture Synthesis by Non-Parametric Sampling}. In
  \bibinfo{booktitle}{{\em Proceedings of the International Conference on
  Computer Vision (ICCV)}} {\em (\bibinfo{series}{ICCV '99})}.
  \bibinfo{publisher}{IEEE Computer Society}, \bibinfo{address}{Washington, DC,
  USA}, \bibinfo{pages}{1033--1038}.
\newblock
\showISBNx{0-7695-0164-8}


\bibitem[\protect\citeauthoryear{Gatys, Ecker, and Bethge}{Gatys
  et~al\mbox{.}}{2016}]%
        {gatys16}
\bibfield{author}{\bibinfo{person}{Leon~A Gatys}, \bibinfo{person}{Alexander~S
  Ecker}, {and} \bibinfo{person}{Matthias Bethge}.}
  \bibinfo{year}{2016}\natexlab{}.
\newblock \showarticletitle{Image style transfer using convolutional neural
  networks}. In \bibinfo{booktitle}{{\em Proceedings of the IEEE Conference on
  Computer Vision and Pattern Recognition (CVPR)}}.
  \bibinfo{pages}{2414--2423}.
\newblock


\bibitem[\protect\citeauthoryear{Gatys, Ecker, Bethge, Hertzmann, and
  Shechtman}{Gatys et~al\mbox{.}}{2017}]%
        {gatys17}
\bibfield{author}{\bibinfo{person}{L.~A. Gatys}, \bibinfo{person}{A.~S. Ecker},
  \bibinfo{person}{M. Bethge}, \bibinfo{person}{A. Hertzmann}, {and}
  \bibinfo{person}{E. Shechtman}.} \bibinfo{year}{2017}\natexlab{}.
\newblock \showarticletitle{Controlling Perceptual Factors in Neural Style
  Transfer}. In \bibinfo{booktitle}{{\em Proceedings of the IEEE Conference on
  Computer Vision and Pattern Recognition (CVPR)}}.
\newblock


\bibitem[\protect\citeauthoryear{Goodfellow, Pouget-Abadie, Mirza, Xu,
  Warde-Farley, Ozair, Courville, and Bengio}{Goodfellow et~al\mbox{.}}{2014}]%
        {gan}
\bibfield{author}{\bibinfo{person}{Ian Goodfellow}, \bibinfo{person}{Jean
  Pouget-Abadie}, \bibinfo{person}{Mehdi Mirza}, \bibinfo{person}{Bing Xu},
  \bibinfo{person}{David Warde-Farley}, \bibinfo{person}{Sherjil Ozair},
  \bibinfo{person}{Aaron Courville}, {and} \bibinfo{person}{Yoshua Bengio}.}
  \bibinfo{year}{2014}\natexlab{}.
\newblock \showarticletitle{Generative adversarial nets}. In
  \bibinfo{booktitle}{{\em Proceedings of Advances in neural information
  processing systems (NIPS)}}. \bibinfo{pages}{2672--2680}.
\newblock


\bibitem[\protect\citeauthoryear{Hedley and Yan}{Hedley and Yan}{1992}]%
        {imageseg}
\bibfield{author}{\bibinfo{person}{Mark Hedley} {and} \bibinfo{person}{Hong
  Yan}.} \bibinfo{year}{1992}\natexlab{}.
\newblock \showarticletitle{Segmentation of color images using spatial and
  color space information.}
\newblock \bibinfo{journal}{{\em J. Electronic Imaging\/}} \bibinfo{volume}{1},
  \bibinfo{number}{4} (\bibinfo{year}{1992}), \bibinfo{pages}{374--380}.
\newblock


\bibitem[\protect\citeauthoryear{Heeger and Bergen}{Heeger and Bergen}{1995}]%
        {pyramid}
\bibfield{author}{\bibinfo{person}{David~J Heeger} {and}
  \bibinfo{person}{James~R Bergen}.} \bibinfo{year}{1995}\natexlab{}.
\newblock \showarticletitle{Pyramid-based texture analysis/synthesis}. In
  \bibinfo{booktitle}{{\em Proceedings of the 22nd annual conference on
  Computer graphics and interactive techniques}}. ACM,
  \bibinfo{pages}{229--238}.
\newblock


\bibitem[\protect\citeauthoryear{Johnson, Alahi, and Fei{-}Fei}{Johnson
  et~al\mbox{.}}{2016}]%
        {faststyle}
\bibfield{author}{\bibinfo{person}{Justin Johnson}, \bibinfo{person}{Alexandre
  Alahi}, {and} \bibinfo{person}{Li Fei{-}Fei}.}
  \bibinfo{year}{2016}\natexlab{}.
\newblock \showarticletitle{Perceptual Losses for Real-Time Style Transfer and
  Super-Resolution}. In \bibinfo{booktitle}{{\em Proceedings of the European
  Conference on Computer Vision ({ECCV})}}. \bibinfo{pages}{694--711}.
\newblock


\bibitem[\protect\citeauthoryear{Kingma and Ba}{Kingma and Ba}{2015}]%
        {adam}
\bibfield{author}{\bibinfo{person}{Diederik Kingma} {and}
  \bibinfo{person}{Jimmy Ba}.} \bibinfo{year}{2015}\natexlab{}.
\newblock \showarticletitle{Adam: A Method for Stochastic Optimization}. In
  \bibinfo{booktitle}{{\em Proceedings of the 3rd International Conference on
  Learning Representations (ICLR)}}.
\newblock


\bibitem[\protect\citeauthoryear{Kwatra, Essa, Bobick, and Kwatra}{Kwatra
  et~al\mbox{.}}{2005}]%
        {example}
\bibfield{author}{\bibinfo{person}{Vivek Kwatra}, \bibinfo{person}{Irfan Essa},
  \bibinfo{person}{Aaron Bobick}, {and} \bibinfo{person}{Nipun Kwatra}.}
  \bibinfo{year}{2005}\natexlab{}.
\newblock \showarticletitle{Texture Optimization for Example-based Synthesis}.
  In \bibinfo{booktitle}{{\em ACM SIGGRAPH}} {\em (\bibinfo{series}{SIGGRAPH
  '05})}. \bibinfo{publisher}{ACM}, \bibinfo{address}{New York, NY, USA},
  \bibinfo{pages}{795--802}.
\newblock


\bibitem[\protect\citeauthoryear{Kwatra, Sch{\"o}dl, Essa, Turk, and
  Bobick}{Kwatra et~al\mbox{.}}{2003}]%
        {graphcut}
\bibfield{author}{\bibinfo{person}{Vivek Kwatra}, \bibinfo{person}{Arno
  Sch{\"o}dl}, \bibinfo{person}{Irfan Essa}, \bibinfo{person}{Greg Turk}, {and}
  \bibinfo{person}{Aaron Bobick}.} \bibinfo{year}{2003}\natexlab{}.
\newblock \showarticletitle{Graphcut textures: image and video synthesis using
  graph cuts}.
\newblock \bibinfo{journal}{{\em ACM Transactions on Graphics (ToG)\/}}
  \bibinfo{volume}{22}, \bibinfo{number}{3} (\bibinfo{year}{2003}),
  \bibinfo{pages}{277--286}.
\newblock


\bibitem[\protect\citeauthoryear{Lee, Choi, and Kim}{Lee et~al\mbox{.}}{2016}]%
        {lap-patch}
\bibfield{author}{\bibinfo{person}{J.~H. Lee}, \bibinfo{person}{I. Choi}, {and}
  \bibinfo{person}{M.~H. Kim}.} \bibinfo{year}{2016}\natexlab{}.
\newblock \showarticletitle{Laplacian Patch-Based Image Synthesis}. In
  \bibinfo{booktitle}{{\em Proceedings of the IEEE Conference on Computer
  Vision and Pattern Recognition (CVPR)}}. \bibinfo{pages}{2727--2735}.
\newblock


\bibitem[\protect\citeauthoryear{Levin, Lischinski, and Weiss}{Levin
  et~al\mbox{.}}{2008}]%
        {matt}
\bibfield{author}{\bibinfo{person}{Anat Levin}, \bibinfo{person}{Dani
  Lischinski}, {and} \bibinfo{person}{Yair Weiss}.}
  \bibinfo{year}{2008}\natexlab{}.
\newblock \showarticletitle{A closed-form solution to natural image matting}.
\newblock \bibinfo{journal}{{\em IEEE Transactions on Pattern Analysis and
  Machine Intelligence (TPAMI)\/}} \bibinfo{volume}{30}, \bibinfo{number}{2}
  (\bibinfo{year}{2008}), \bibinfo{pages}{228--242}.
\newblock


\bibitem[\protect\citeauthoryear{Li and Wand}{Li and Wand}{2016}]%
        {mrf}
\bibfield{author}{\bibinfo{person}{Chuan Li} {and} \bibinfo{person}{Michael
  Wand}.} \bibinfo{year}{2016}\natexlab{}.
\newblock \showarticletitle{Combining markov random fields and convolutional
  neural networks for image synthesis}. In \bibinfo{booktitle}{{\em Proceedings
  of the IEEE Conference on Computer Vision and Pattern Recognition (CVPR)}}.
  \bibinfo{pages}{2479--2486}.
\newblock


\bibitem[\protect\citeauthoryear{Luan, Paris, Shechtman, and Bala}{Luan
  et~al\mbox{.}}{2017}]%
        {photo}
\bibfield{author}{\bibinfo{person}{Fujun Luan}, \bibinfo{person}{Sylvain
  Paris}, \bibinfo{person}{Eli Shechtman}, {and} \bibinfo{person}{Kavita
  Bala}.} \bibinfo{year}{2017}\natexlab{}.
\newblock \showarticletitle{Deep Photo Style Transfer}. In
  \bibinfo{booktitle}{{\em Proceedings of the IEEE Conference on Computer
  Vision and Pattern Recognition (CVPR)}}.
\newblock


\bibitem[\protect\citeauthoryear{Nie, Wang, Zha, Li, and Chua}{Nie
  et~al\mbox{.}}{2011}]%
        {Nie:2011SIGIR}
\bibfield{author}{\bibinfo{person}{Liqiang Nie}, \bibinfo{person}{Meng Wang},
  \bibinfo{person}{Zhengjun Zha}, \bibinfo{person}{Guangda Li}, {and}
  \bibinfo{person}{Tat-Seng Chua}.} \bibinfo{year}{2011}\natexlab{}.
\newblock \showarticletitle{Multimedia Answering: Enriching Text QA with Media
  Information}. In \bibinfo{booktitle}{{\em Proceedings of the 34th
  International ACM SIGIR Conference on Research and Development in Information
  Retrieval}} {\em (\bibinfo{series}{SIGIR '11})}. \bibinfo{publisher}{ACM},
  \bibinfo{pages}{695--704}.
\newblock
\showISBNx{978-1-4503-0757-4}


\bibitem[\protect\citeauthoryear{Nie, Yan, Wang, Hong, and Chua}{Nie
  et~al\mbox{.}}{2012}]%
        {Nie:2012SIGMM}
\bibfield{author}{\bibinfo{person}{Liqiang Nie}, \bibinfo{person}{Shuicheng
  Yan}, \bibinfo{person}{Meng Wang}, \bibinfo{person}{Richang Hong}, {and}
  \bibinfo{person}{Tat-Seng Chua}.} \bibinfo{year}{2012}\natexlab{}.
\newblock \showarticletitle{Harvesting Visual Concepts for Image Search with
  Complex Queries}. In \bibinfo{booktitle}{{\em Proceedings of the 20th ACM
  International Conference on Multimedia (MM)}} {\em
  (\bibinfo{series}{MM'12})}. \bibinfo{publisher}{ACM},
  \bibinfo{pages}{59--68}.
\newblock
\showISBNx{978-1-4503-1089-5}


\bibitem[\protect\citeauthoryear{{Nikulin} and {Novak}}{{Nikulin} and
  {Novak}}{2016}]%
        {explore}
\bibfield{author}{\bibinfo{person}{Y. {Nikulin}} {and} \bibinfo{person}{R.
  {Novak}}.} \bibinfo{year}{2016}\natexlab{}.
\newblock \showarticletitle{{Exploring the Neural Algorithm of Artistic
  Style}}.
\newblock \bibinfo{journal}{{\em ArXiv e-prints\/}} (\bibinfo{date}{Feb.}
  \bibinfo{year}{2016}).
\newblock
\showeprint[arxiv]{cs.CV/1602.07188}


\bibitem[\protect\citeauthoryear{P{\'e}rez, Gangnet, and Blake}{P{\'e}rez
  et~al\mbox{.}}{2003}]%
        {poisson}
\bibfield{author}{\bibinfo{person}{Patrick P{\'e}rez}, \bibinfo{person}{Michel
  Gangnet}, {and} \bibinfo{person}{Andrew Blake}.}
  \bibinfo{year}{2003}\natexlab{}.
\newblock \showarticletitle{Poisson Image Editing}. In \bibinfo{booktitle}{{\em
  ACM SIGGRAPH}} {\em (\bibinfo{series}{SIGGRAPH '03})}.
  \bibinfo{publisher}{ACM}, \bibinfo{address}{New York, NY, USA},
  \bibinfo{pages}{313--318}.
\newblock
\showISBNx{1-58113-709-5}


\bibitem[\protect\citeauthoryear{Portilla and Simoncelli}{Portilla and
  Simoncelli}{2000}]%
        {wavelet}
\bibfield{author}{\bibinfo{person}{Javier Portilla} {and}
  \bibinfo{person}{Eero~P Simoncelli}.} \bibinfo{year}{2000}\natexlab{}.
\newblock \showarticletitle{A parametric texture model based on joint
  statistics of complex wavelet coefficients}.
\newblock \bibinfo{journal}{{\em International Journal of Computer Vision\/}}
  \bibinfo{volume}{40}, \bibinfo{number}{1} (\bibinfo{year}{2000}),
  \bibinfo{pages}{49--70}.
\newblock


\bibitem[\protect\citeauthoryear{Prince}{Prince}{2012}]%
        {cv-prince}
\bibfield{author}{\bibinfo{person}{Simon~JD Prince}.}
  \bibinfo{year}{2012}\natexlab{}.
\newblock \bibinfo{booktitle}{{\em Computer vision: models, learning, and
  inference}}.
\newblock \bibinfo{publisher}{Cambridge University Press}.
\newblock


\bibitem[\protect\citeauthoryear{Wei and Levoy}{Wei and Levoy}{2000}]%
        {tree}
\bibfield{author}{\bibinfo{person}{Li-Yi Wei} {and} \bibinfo{person}{Marc
  Levoy}.} \bibinfo{year}{2000}\natexlab{}.
\newblock \showarticletitle{Fast texture synthesis using tree-structured vector
  quantization}. In \bibinfo{booktitle}{{\em Proceedings of the 27th annual
  conference on Computer graphics and interactive techniques}}. ACM
  Press/Addison-Wesley Publishing Co., \bibinfo{pages}{479--488}.
\newblock


\bibitem[\protect\citeauthoryear{Zhang, Kyaw, Chang, and Chua}{Zhang
  et~al\mbox{.}}{2017}]%
        {zhang2017relation}
\bibfield{author}{\bibinfo{person}{Hanwang Zhang}, \bibinfo{person}{Zawlin
  Kyaw}, \bibinfo{person}{Shih-Fu Chang}, {and} \bibinfo{person}{Tat-Seng
  Chua}.} \bibinfo{year}{2017}\natexlab{}.
\newblock \showarticletitle{Visual Translation Embedding Network for Visual
  Relation Detection}. In \bibinfo{booktitle}{{\em Proceedings of the IEEE
  Conference on Computer Vision and Pattern Recognition (CVPR)}}.
\newblock


\bibitem[\protect\citeauthoryear{Zhang, Shang, Luan, Wang, and Chua}{Zhang
  et~al\mbox{.}}{2016}]%
        {zhang2016learning}
\bibfield{author}{\bibinfo{person}{Hanwang Zhang}, \bibinfo{person}{Xindi
  Shang}, \bibinfo{person}{Huanbo Luan}, \bibinfo{person}{Meng Wang}, {and}
  \bibinfo{person}{Tat-Seng Chua}.} \bibinfo{year}{2016}\natexlab{}.
\newblock \showarticletitle{Learning from collective intelligence: Feature
  learning using social images and tags}.
\newblock \bibinfo{journal}{{\em ACM Transactions on Multimedia Computing,
  Communications, and Applications (TOMM)\/}}  \bibinfo{volume}{13}
  (\bibinfo{year}{2016}).
\newblock


\bibitem[\protect\citeauthoryear{{Zhu}, {Park}, {Isola}, and {Efros}}{{Zhu}
  et~al\mbox{.}}{2017}]%
        {cyclegan}
\bibfield{author}{\bibinfo{person}{J.-Y. {Zhu}}, \bibinfo{person}{T. {Park}},
  \bibinfo{person}{P. {Isola}}, {and} \bibinfo{person}{A.~A. {Efros}}.}
  \bibinfo{year}{2017}\natexlab{}.
\newblock \showarticletitle{{Unpaired Image-to-Image Translation using
  Cycle-Consistent Adversarial Networks}}. In \bibinfo{booktitle}{{\em
  Proceedings of the IEEE International Conference on Computer Vision (ICCV)}}.
\newblock


\end{thebibliography}

\end{document}